\documentclass[10pt,twocolumn,letterpaper]{article}
\usepackage[final]{cvpr}
\usepackage{hyperref}

\title{Exploring Prompt Alignment with Clinical Factors in Zero-Shot Segmentation VLMs for NSCLC Tumor Segmentation}

\author{
Suraj Pai$^{1}$ \quad Thibault Heintz$^{1}$ \quad Cosmin Ciausu$^{1}$\\
Marion Tonneau$^{1}$ \quad Hugo Aerts$^{1}$ \quad Raymond Mak$^{1}$\\
$^{1}$Artificial Intelligence in Medicine Program,\\
Mass General Brigham, Harvard Medical School, USA
}

\begin{document}
\maketitle

\begin{abstract}
  Zero-shot vision-language models (VLMs) offer a promptable alternative to task-specific training for gross tumor volume (GTV) delineation in non-small-cell lung cancer (NSCLC), but the prompt dimensions that govern their spatial behavior remain poorly understood. We study this question by probing alignment directions in VoxTell on a held-out internal NSCLC tumor dataset through sub-prompt decomposition into diagnosis, demographic, staging, anatomical, generic, and irrelevant controls; attribute-wise perturbation robustness; specificity ladders; and cross-case prompt swaps, while benchmarking against fine-tuned and zero-shot baselines using the Dice Similarity Coefficient (DSC) with Wilcoxon signed-rank tests and Benjamini--Hochberg correction.
  Alignment analyses revealed that anatomical location is the dominant driver of VoxTell's spatial attention: 63.4\% of location perturbations caused catastrophic drops, prompt specificity improved from generic to full descriptions except for diagnosis-only prompts, irrelevant prompts correctly yielded zero segmentation, and cross-case prompt swaps confirmed patient-specific conditioning (matched DSC 0.906 vs.\ mismatched 0.406). Histology and stage substitutions had minimal effect, indicating that the model prioritizes ``where to look'' over ``what to look for.''
  In this context, VoxTell, operating fully zero-shot, achieved a mean DSC of 0.613, statistically indistinguishable from nnUNet (0.690, $p_\text{adj}=0.156$) and Ahmed et al.~\cite{hosny2022clinical} (0.675, $p_\text{adj}=0.679$), while significantly outperforming all other zero-shot models. Together, these findings argue that segmentation VLMs should be evaluated not only by Dice, but also by the prompt dimensions to which they align.
\end{abstract}

\section{Introduction}

Accurate GTV delineation is a prerequisite for safe radiotherapy planning in NSCLC. Manual contouring requires 30--90 minutes per patient and exhibits DSC disagreements of 10--30\% between expert annotators~\cite{guzene2023interobserver}. Deep learning can reduce contouring time by 65\% while preserving dosimetric equivalence~\cite{hosny2022clinical}, and automated segmentation has achieved DSC values up to 0.82 under controlled conditions~\cite{primakov2022nsclc}. For promptable zero-shot VLMs, however, benchmark performance alone is insufficient: clinical use also depends on understanding which prompt attributes actually drive the predicted tumor mask, which attributes can be safely ignored, and whether failure modes remain clinically interpretable.

The landscape of segmentation models now spans three paradigms: \textit{fine-tuned supervised} models such as nnUNet~\cite{isensee2021nnunet}; \textit{zero-shot foundation} models including TotalSegmentator~\cite{wasserthal2023totalsegmentator}, VISTA3D~\cite{he2024vista3d}, BiomedParse~\cite{zhao2024biomedparse}, SAT~\cite{zhao2023sat}, and CAT~\cite{huang2024cat}; and \textit{zero-shot vision-language models} (VLMs) that accept both volumetric imaging and free-text clinical prompts. However, no prior study has compared all three paradigms on a clinically validated, independently collected dataset while also characterizing which clinical prompt dimensions a zero-shot VLM aligns with during segmentation. This question is especially important in radiotherapy, where prompts may contain diagnosis, TNM stage, demographics, laterality, and anatomic descriptors, yet only some of these attributes should influence the primary tumor mask. Recent work confirms that foundation model performance on lung tumor segmentation varies substantially across evaluation settings~\cite{mulero2025benchmark}.

A clinically useful segmentation VLM should therefore satisfy at least three properties: it should improve as prompts become more spatially informative, suppress output for irrelevant prompts, and fail under mismatched case information rather than defaulting to a generic tumor detector. We address this gap by probing alignment directions in the VLM VoxTell on cases from an internal NSCLC tumor dataset (HarvardRT)~\cite{hosny2022clinical} through sub-prompt decomposition, perturbation robustness, specificity ladders, and cross-case prompt swaps, and by situating these findings within a benchmark of nine models spanning fine-tuned supervised, zero-shot foundation, and zero-shot VLM paradigms.

\section{Methods}

\textbf{Dataset.}
The held-out HarvardRT set comprises 93 NSCLC cases, each with a planning CT and expert GTV contour. Free-text clinical prompts were constructed from institutional clinical notes for text-accepting models and preserved diagnosis, demographic, and staging language when available, enabling controlled decomposition and attribute-level perturbation.

\textbf{Alignment analyses.}
For the best-performing zero-shot VLM (VoxTell), we conducted four experiments on a seven-case subset spanning diverse histologies and locations; the cross-case swap used a representative five-case subset to form a $5\times5$ design. In \textit{sub-prompt fragment analysis}, each original prompt was decomposed into fragments such as diagnosis, demographics, TNM, stage, diagnosis+stage, generic (``lung tumor''), anatomical, and irrelevant (``liver cyst'') controls. In \textit{perturbation robustness}, we preserved sentence structure but swapped one attribute at a time, including tumor type, overall stage, T/N/M stage, age, sex, and special control prompts, and measured $\Delta$DSC relative to the matched prompt. In the \textit{specificity ladder}, we varied prompt detail from L0 (``tumor'') to L6 (full prompt plus fabricated detail) to test whether additional clinical detail improves or destabilizes alignment. In \textit{cross-case prompt swaps}, each image was paired with prompts from other patients to distinguish patient-specific conditioning from prompt-agnostic tumor detection. Primary readouts were DSC, $\Delta$DSC, zero-mask failure rates, and matched-versus-mismatched behavior.

\textbf{Models.}
Nine models were evaluated: \textit{fine-tuned}---nnUNet (GTV-only), Ahmed et al.\ (multi-stage) ~\cite{hosny2022clinical}, MM-LLM-RO (Gemma-3 4B)~\cite{oh2024mmllmro}; \textit{zero-shot vision (ZS-Vision)}---TotalSeg-Nodules, VISTA3D; \textit{zero-shot VLM (ZS-VLM)}---VoxTell~\cite{voxtell2024}, BiomedParse~\cite{zhao2024biomedparse}, SAT~\cite{zhao2023sat}, CAT~\cite{huang2024cat}. For models with multiple prompt variants, the best-performing variant was selected.

\textbf{Benchmark evaluation.}
DSC was the primary metric. A Friedman test assessed overall heterogeneity; pairwise Wilcoxon signed-rank tests with Benjamini--Hochberg FDR correction ($\alpha=0.05$) were used for all 33 pairwise comparisons.

\section{Results}

\textbf{Alignment results.}
The fragment analysis showed full clinical prompts achieve the highest DSC ($0.861\pm0.081$), while generic prompts (``lung tumor'') still produce reasonable output ($0.723\pm0.269$) and irrelevant prompts (``liver cyst'') correctly yield zero segmentation. Diagnosis-only fragments were inconsistent ($0.653\pm0.312$), performing well for common histologies (e.g., ``adenocarcinoma,'' DSC=0.934) but failing for rare types (e.g., ``leiomyosarcoma,'' DSC=0.000). This gap between generic and full prompts indicates that VoxTell is not simply acting as a prompt-agnostic tumor trigger, yet the variability of diagnosis-only prompts suggests that disease labels alone are insufficient to anchor the mask reliably.

\begin{figure}[t]
  \centering
  \includegraphics[width=\columnwidth]{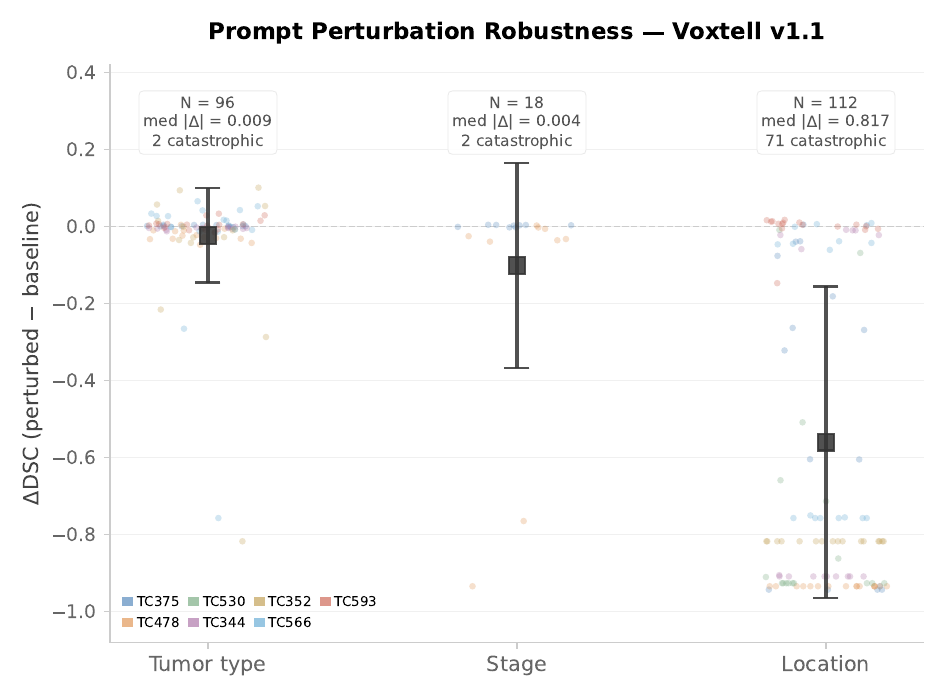}
  \caption{Prompt perturbation robustness. $\Delta$DSC distribution by perturbation category. Location swaps cause catastrophic degradation (63.4\% with $|\Delta\text{DSC}|>0.5$); tumor-type and stage swaps are largely benign.}
  \label{fig:perturb}
\end{figure}

Perturbation experiments revealed a striking asymmetry (Figure~\ref{fig:perturb}): tumor-type swaps had minimal effect (median $|\Delta\text{DSC}|=0.009$, 2/96 catastrophic), but 63.4\% of location swaps caused catastrophic drops ($|\Delta\text{DSC}|>0.5$, mean $\Delta\text{DSC}=-0.560$). Changing tumor type, stage, or demographics typically produced only modest shifts, whereas location edits often displaced the predicted mask or eliminated it entirely. This selective sensitivity indicates that the model disentangles spatial localization from diagnostic labels and is most strongly aligned to prompt components with direct spatial implications.

\begin{figure}[t]
  \centering
  \includegraphics[width=\columnwidth]{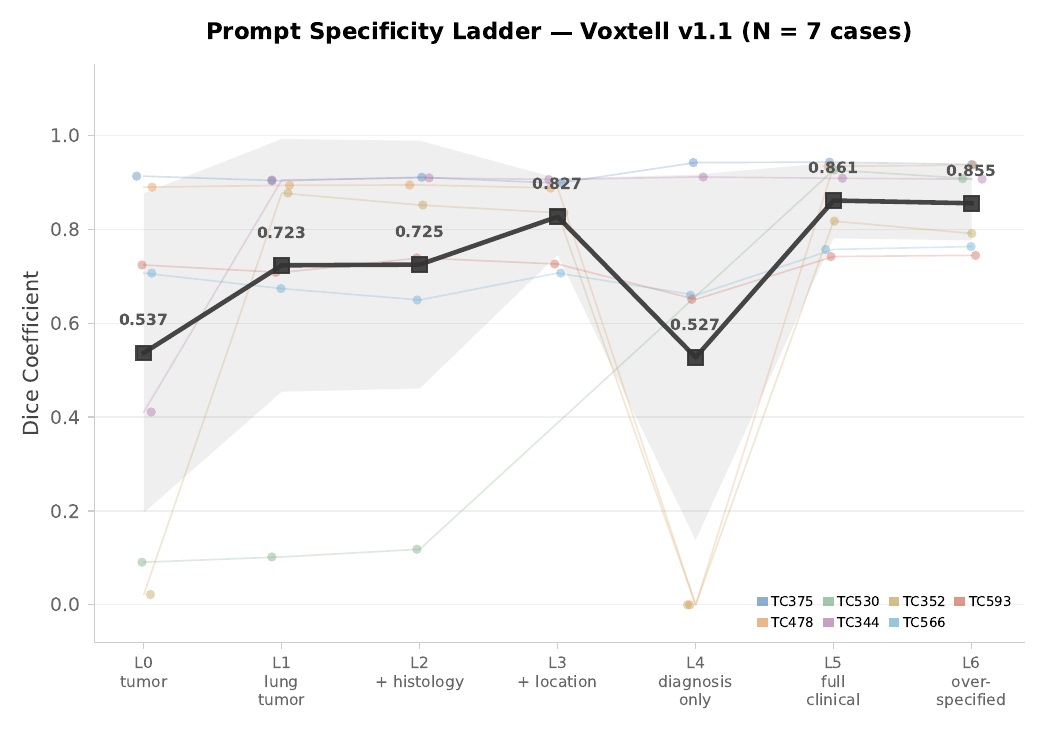}
  \caption{Prompt specificity ladder. DSC shows near-monotonic improvement from L0 (``tumor'') to L5 (full clinical prompt), with an exception at diagnosis-only L4. Largest gains occur at the organ-to-lobe specificity transition. Over-specification (L6) provides no benefit.}
  \label{fig:ladder}
\end{figure}

The specificity ladder (Figure~\ref{fig:ladder}) showed near-monotonic improvement from L0 (``tumor'', DSC=0.537) through L1 (``lung tumor'', 0.723) to L5 (full clinical prompt, 0.861), except for diagnosis-only at L4, with the largest gains at the transition from organ-level to lobe-level specificity. The large jump from L0 to L1 indicates that organ-level context provides the first major gain over a generic query, while the dip at L4 suggests that structured diagnosis-and-staging tokens without naturalistic anatomic context are less useful than a coherent free-text description. Over-specification (L6) offered no additional benefit (0.855), suggesting that fabricated detail can saturate or slightly confuse the alignment signal.

\begin{figure}[t]
  \centering
  \includegraphics[width=\columnwidth]{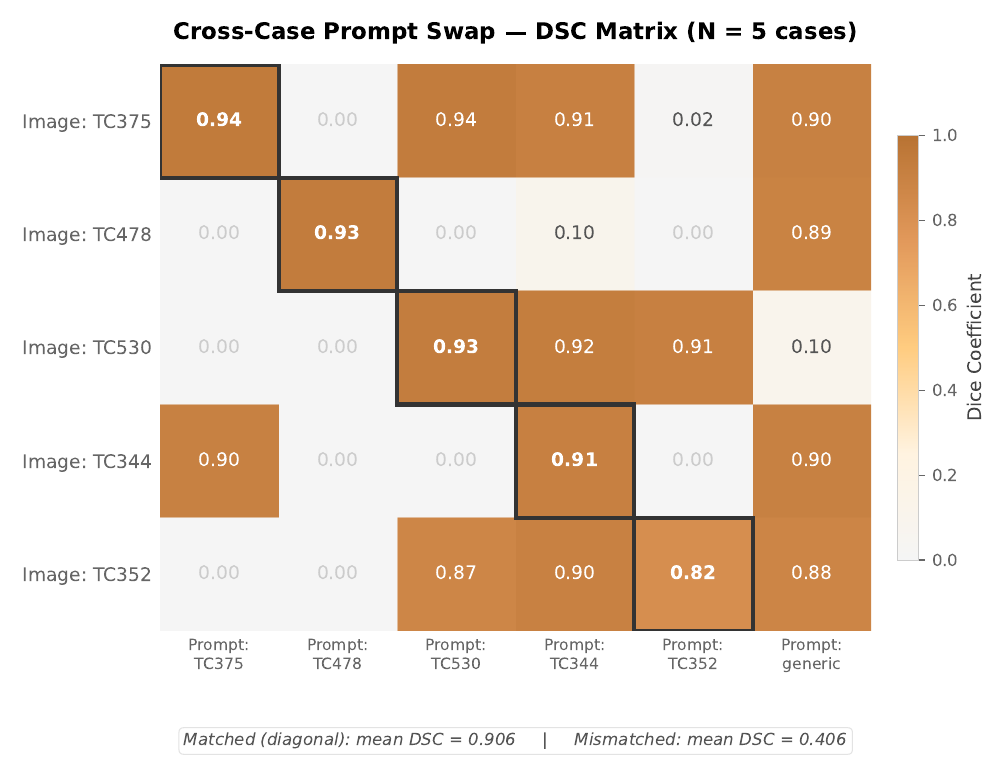}
  \caption{Cross-case prompt swap: $5\times5$ DSC matrix. Diagonal (matched) entries substantially outperform off-diagonal (mismatched) entries. Generic ``lung tumor'' prompt shown in rightmost column.}
  \label{fig:heatmap}
\end{figure}

Cross-case prompt swaps confirmed patient-specific conditioning: matched pairs achieved DSC $0.906\pm0.046$ vs.\ $0.406\pm0.441$ for mismatched pairs, with 44\% of mismatches producing zero output (Figure~\ref{fig:heatmap}). The strong diagonal dominance of the swap matrix argues against a generic ``any tumor'' behavior and instead suggests that VoxTell uses prompt content to determine both where to segment and whether a lesion should be segmented on that scan at all.

\textbf{Overall performance.}
To contextualize these alignment findings, Table~\ref{tab:results} summarizes model performance. A Friedman test confirmed significant heterogeneity ($\chi^2=318.2$, $p<0.001$). nnUNet (0.690) and Ahmed et al.~\cite{hosny2022clinical} (0.675) led overall; VoxTell (0.613) ranked fourth; most zero-shot VLM models performed poorly (DSC $<0.26$). Within the zero-shot setting, however, VoxTell clearly separated itself from the other text-conditioned models, indicating that competitive segmentation quality can coexist with interpretable prompt dependence rather than requiring prompt invariance.

\begin{table}[t]
  \centering
  \caption{GTV segmentation performance on HarvardRT (93 cases).}
  \label{tab:results}
  \small
  \setlength{\tabcolsep}{3pt}
  \begin{tabular}{@{}llcc@{}}
    \toprule
    Model & Category & Mean DSC & Median \\
    \midrule
    nnUNet (GTV) & Fine-tuned & $0.690\pm0.272$ & 0.810 \\
    Ahmed et al. & Fine-tuned & $0.675\pm0.269$ & 0.801 \\
    MM-LLM-RO & Fine-tuned & $0.434\pm0.280$ & 0.448 \\
    TotalSeg-Nod. & ZS-Vision & $0.653\pm0.292$ & 0.775 \\
    VISTA3D & ZS-Vision & $0.417\pm0.280$ & 0.442 \\
    VoxTell & ZS-VLM & $0.613\pm0.348$ & 0.777 \\
    BiomedParse & ZS-VLM & $0.252\pm0.247$ & 0.141 \\
    SAT & ZS-VLM & $0.225\pm0.281$ & 0.059 \\
    CAT & ZS-VLM & $0.158\pm0.316$ & 0.000 \\
    \bottomrule
  \end{tabular}
\end{table}

\begin{figure}[t]
  \centering
  \includegraphics[width=\columnwidth]{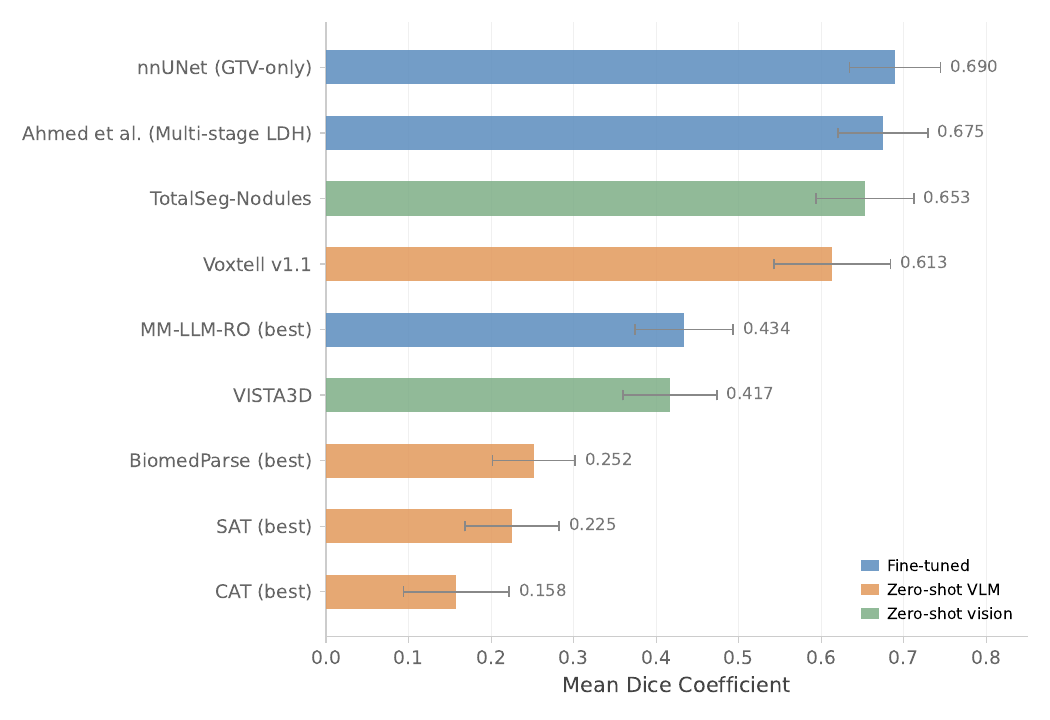}
  \caption{Per-case DSC distributions for all nine models. VoxTell (orange, zero-shot VLM) shows comparable central tendency to the fine-tuned leaders (blue), though with wider spread.}
  \label{fig:boxplot}
\end{figure}

\begin{figure}[t]
  \centering
  \includegraphics[width=\columnwidth]{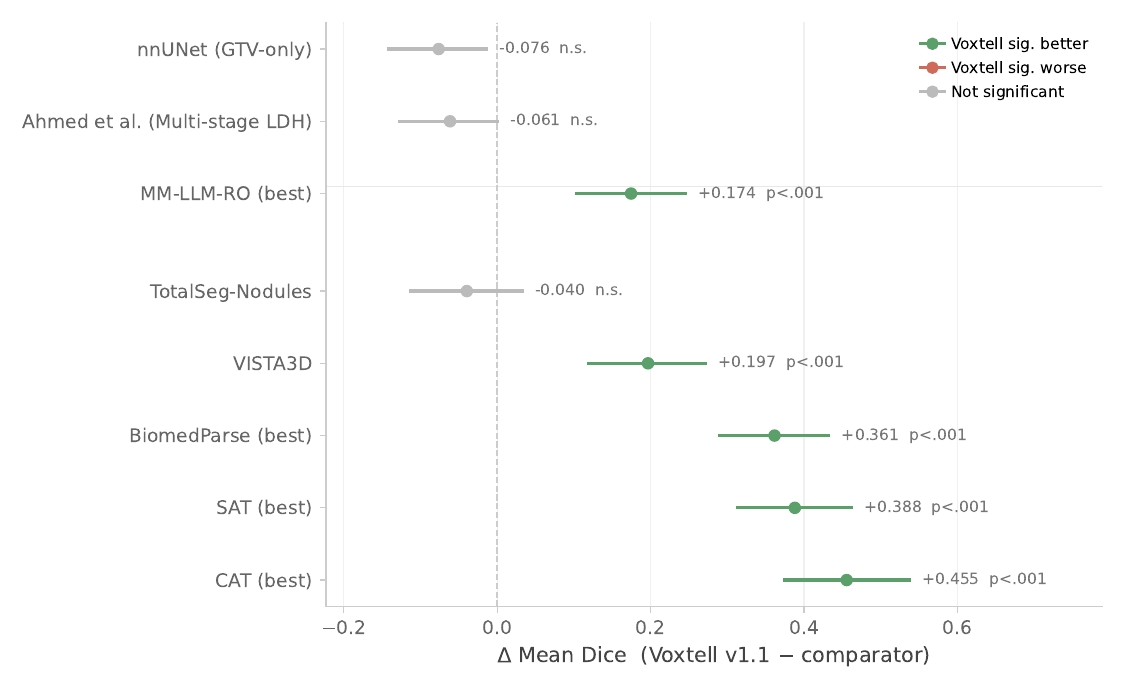}
  \caption{Pairwise DSC differences relative to VoxTell. Positive values indicate VoxTell outperformance. Significance after BH correction ($\alpha=0.05$).}
  \label{fig:forest}
\end{figure}

\textbf{VoxTell vs.\ fine-tuned models.}
VoxTell did not differ significantly from nnUNet ($p_\text{adj}=0.156$, $r=0.216$) or Ahmed et al.~\cite{hosny2022clinical} ($p_\text{adj}=0.679$, $r=0.098$). It significantly outperformed MM-LLM-RO ($p_\text{adj}<0.001$, $r=0.607$) and all zero-shot foundation models except TotalSeg-Nodules ($p_\text{adj}=0.982$). Per-case DSC distributions are shown in Figure~\ref{fig:boxplot} and pairwise comparisons in Figure~\ref{fig:forest}. This benchmark result is important because it shows that strong prompt alignment does not come at the cost of gross segmentation accuracy.

\section{Discussion and Conclusion}

This study shows that alignment in a zero-shot segmentation VLM is not distributed uniformly across clinical prompt content. In VoxTell, anatomical location is the dominant alignment direction, while histology and stage contribute far less, even as the model achieves performance statistically indistinguishable from the best fine-tuned baselines on an independent clinical dataset. From a clinical interface perspective, this means that promptability is meaningful, but primarily for the subset of prompt content that carries actionable spatial information.

Placed in benchmark context, several factors explain the performance hierarchy. nnUNet and Ahmed et al.~\cite{hosny2022clinical} benefit from direct exposure to the HarvardRT training distribution, while TotalSeg-Nodules transfers well due to visual overlap between lung nodules and GTVs. The poor performance of BiomedParse, SAT, and CAT (DSC $<$ 0.26) likely reflects a mismatch between their multi-organ training distributions and the heterogeneous appearance of lung tumors, consistent with recent findings that foundation model performance varies substantially across evaluation settings~\cite{mulero2025benchmark}. MM-LLM-RO's underperformance despite fine-tuning (0.434 vs.\ 0.690 for nnUNet) suggests that current multimodal LLM architectures introduce optimization challenges that offset the advantages of incorporating clinical text. Notably, VoxTell is the only zero-shot VLM in this comparison that approaches the fine-tuned frontier while also exhibiting clear prompt-conditioned behavior.

Viewed through this alignment lens, VoxTell leverages clinical text in a selective rather than uniform manner. The model disentangles spatial localization cues from diagnostic labels, prioritizing ``where to look'' over ``what to look for'': anatomical location is the single most influential prompt component, while histological subtype has negligible effect on segmentation quality. The specificity ladder shows near-monotonic improvement with increasing prompt detail except for diagnosis-only at L4, irrelevant prompts are correctly suppressed, and the cross-case swap experiment confirms that the model produces genuinely patient-specific segmentations conditioned on prompt content rather than generic tumor masks. This pattern is consistent with architectures in which text queries first localize regions through cross-attention and are then fused with spatial decoder features by dot-product operations, making anatomy-bearing tokens more immediately actionable than abstract descriptors such as stage or histology. These findings suggest that VLMs ground clinical language in volumetric spatial representations along interpretable alignment directions---a capability that distinguishes them from both conventional supervised models and generic foundation models.

\textbf{Limitations.}
Evaluation was conducted at a single institution (93 cases); DSC was the sole metric; alignment analyses used only seven cases; and automated prompt optimization was not explored. We also do not exhaustively report auxiliary quantities such as logit contrast or embedding-space geometry in the main text, focusing instead on segmentation behavior that is easiest to interpret clinically. Notably, VoxTell's segmentation is driven almost entirely by anatomical location, with histology and stage having negligible effect. While sufficient for primary tumor delineation, this anatomical bias may limit applicability to more complex targets such as nodal volumes, where clinical staging, imaging findings, and treatment intent must be integrated to determine which structures to contour.

\textbf{Conclusion.}
The dominant alignment direction in zero-shot VLM tumor segmentation is anatomical location: VoxTell responds strongly to spatial prompt cues, suppresses irrelevant prompts, and produces patient-specific segmentations under matched clinical conditioning. In benchmark context, it also approaches fine-tuned supervised performance for GTV segmentation. Together, these results suggest that evaluation of segmentation VLMs should include both benchmark accuracy and prompt-alignment analyses. More broadly, integrating clinical language understanding with volumetric image analysis remains a promising direction for scalable radiotherapy contouring, provided future work validates alignment behavior across institutions, targets, and dosimetric endpoints.

\section*{Acknowledgments}
We thank Maximilian Rokuss and Moritz Langenberg, authors of VoxTell, for developing the model evaluated in this work and for their assistance with better understanding the model and framework. We also thank the team at Brigham and Womens Hospital/Dana Farber Cancer Institute, behind the collection, curation and maintenance of the HarvardRT dataset.

{\small
  \bibliographystyle{plainnat}
  \bibliography{references}
}

\end{document}